\documentclass[conference]{IEEEtran}
\IEEEoverridecommandlockouts
\usepackage{cite}
\usepackage{amsmath,amssymb,amsfonts}
\usepackage{algorithmic}
\usepackage{enumitem}
\usepackage{subfigure}
\usepackage{graphicx}
\usepackage{textcomp}
\usepackage{xcolor}
\def\BibTeX{{\rm B\kern-.05em{\sc i\kern-.025em b}\kern-.08em
    T\kern-.1667em\lower.7ex\hbox{E}\kern-.125emX}}
\begin{document}
\definecolor{rev}{RGB}{0,0,0}
\definecolor{revAle}{RGB}{0, 0, 0}

\title{Cloud Failure Prediction with Hierarchical Temporal Memory: An Empirical Assessment\\
}

\author{\IEEEauthorblockN{Oliviero Riganelli\IEEEauthorrefmark{1}, Paolo Saltarel \IEEEauthorrefmark{2}, Alessandro Tundo\IEEEauthorrefmark{1}, Marco Mobilio\IEEEauthorrefmark{1} and Leonardo Mariani\IEEEauthorrefmark{1}}
	\IEEEauthorblockA{\textit{University of Milano - Bicocca}, Milan, Italy \\
	\IEEEauthorrefmark{1} \textit{\{name.surname\}}@unimib.it, \IEEEauthorrefmark{2} p.saltarel@campus.unimib.it}
}

\maketitle

\begin{abstract}
Hierarchical Temporal Memory (HTM) is an unsupervised learning algorithm inspired by the features of the neocortex that can be used to continuously process stream data and detect anomalies, without requiring a large amount of data for training nor requiring labeled data. HTM is also able to continuously learn from samples, providing a model that is always up-to-date with respect to observations. 

These characteristics make HTM particularly suitable for \textcolor{revAle}{supporting} online failure prediction in cloud systems, which are systems with a dynamically changing behavior that must be monitored to anticipate problems. 
This paper presents the first systematic study that assesses HTM in the context of failure prediction. 


The results that we obtained considering $72$ configurations of HTM applied to $12$ different types of faults introduced in the Clearwater cloud system show that HTM can \textcolor{revAle}{help to}  predict failures with sufficient effectiveness (F-measure $= 0.76$), representing an interesting practical alternative to (semi-)supervised algorithms.
\end{abstract}


\section{Introduction} \label{sec:introduction}

Cloud systems are distributed systems that rely on virtualization technologies to flexibly scale services depending on the environmental conditions, such as the workload and the available resources. Their capability to adapt to the conditions that emerge in the field, while they are operating, is extremely useful to save resources and finally deliver the expected quality of service despite changing conditions~\cite{Chen:CloudAutoscaling:CSUR:2018}. 

Unfortunately, the adaptability and flexibility of the cloud sophisticate the environment, and consequently it increases the chance to observe misbehaviors and failures. In fact, in contrast with the strong availability requirements of modern Web applications and services, empirical data show that failures are still extremely frequent in cloud systems~\cite{Chen:GoogleJobs:ISSRE:2014,Lin:PredictingNodeFailures:ESECFSE:2018,Cotroneo:OpenStackFailures:ESECFSE:2019}. For instance, Microsoft reported a failure rate that drastically impacts on the target availability of 99.999\%~\cite{Lin:PredictingNodeFailures:ESECFSE:2018}, and Vishwanath and Nagappan reported frequent sever failures in data centers~\cite{Vishwanath:CloudHwReliability:ESECFSE:2010}. Further, Cotroneo et al. show that many of the failures experienced in the cloud are not timely detected and notified~\cite{Cotroneo:OpenStackFailures:ESECFSE:2019}.

To handle failures properly, cloud systems are equipped with monitoring techniques that collect behavioral data 
about both the individual services and the infrastructure~\cite{amazon_2021_cloudwatch,prometheus,elk},  to detect anomalies, raise alarms, and anticipate failures~\cite{Zhao:RealTimePrediction:ESECFSE:2020,Mariani:PredictingFailures:JSS:2020,Gu:IncidentIdentification:ESECFSE:2020,Lin:PredictingNodeFailures:ESECFSE:2018,Lin:IncidentManagement:KDD:2014}.

In this paper, we focus on the challenge of predicting failure occurrences exploiting runtime data. So far, this challenge has been addressed with machine learning algorithms, such as Support Vector Machine~\cite{Kanoun:AnomalyDetectionVNF:ISSRE:2016}, Long Short Term Memory~\cite{Lin:PredictingNodeFailures:ESECFSE:2018} or Gradient Boosting Tree~\cite{Zhao:RealTimePrediction:ESECFSE:2020}, that require a \emph{supervised training phase that must be frequently repeated} to adapt the models to the changing behavior of the monitored cloud system. 

Hierarchical Temporal Memory (HTM) is a brain-inspired \emph{unsupervised} machine learning algorithm originally proposed by Jeff Hawkins \cite{Hawkins:OnIntelligence:2004}. 
HTM mimics the structural and functional characteristics of cells in the cerebral cortex in order to learn and make predictions, while effectively processing \emph{spatial} and \emph{temporal} information extracted from input data stream. HTM \emph{continually updates} models as new data to analyze become available.
While successfully applied in many other contexts~\cite{HTMMedicalStreams, HTMGazeGestureRecognition,khangamwa2010detecting, zhang2020online,HTMCellularPhoneIntention,HTMTrafficFlow,Bamaqa:HTMCrowd:ICCBDC:20}, it has received little attention in the context of cloud failure prediction, with only a preliminary study by Mobilio et al. considering its usage~\cite{Mobilio:ADaaS:ISSREW:2019}.

This paper presents the 
\textcolor{revAle}{\emph{first study that systematically evaluates the effectiveness of 72 configurations of HTM in supporting cloud failure prediction}}.
The study considers a realistic telecom cloud-based system that provides IP-based voice, video and message services. We evaluated the capability to predict failures using anomalies reported by HTM for 12 types of faults while analyzing Key Performance Indicators (KPIs), such as CPU, network and memory consumption, collected from the cloud resources available in the system. Results show that anomalies reported by HTM can be used to predict failure occurrences with good effectiveness (F-measure $= 0.76$). 


The paper is organized as follows. Section~\ref{sec:background} introduces HTM. Section~\ref{sec:prediction} presents the online failure prediction strategy used to evaluate HTM. Section~\ref{sec:methodology} explains the evaluation methodology. Section~\ref{sec:results} describes results. Section~\ref{sec:related_work} discusses related work. Section~\ref{sec:conclusions} provides final remarks.  

\section{Hierarchical Temporal Memory} \label{sec:background}

Hierarchical Temporal Memory (HTM) is an unsupervised learning algorithm inspired by the structural and algorithmic features of the neocortex, which consists of many structurally-identical regions responsible for different tasks, despite their similar cellular structure. Similar to the neocortex, HTM is composed of regions interconnected in a hierarchy. Each region is able to learn patterns detected from streamed input values. The regions are connected in a hierarchy, so that information can flow across regions.  Such a structure allows high-level representations to be formed from low-level sensing data, mimicking the behavior of the brain that does not store every single object of a class to recognize it, but it can simply memorize the properties that define the objects of that class.

Sparse Distributed Representations (SDRs) are the means of memorizing and transferring information in HTM. An SDR is a large vector of bits of which only a small percentage is active to capture the semantic meaning of the encoded information. 
This implies that if the set of active bits of one SDR overlaps strongly with the set of active bits of another SDR, then those SDRs are semantically similar. 


HTM is composed of two main components, the Spatial Pooler (SP) and the Sequence Memory (SM), running in a pipeline. The SP has the responsibility of enconding the 
input data $x_t$ into a SDR.  
The SP learns the spatial characteristics of each input and finds a stable representation of the spatial patterns in the form of a sparse vector $a(x_t)$. 
For example, the time series data (i.e., sequence of numerical values associated with a timestamp) produced by each KPI can be encoded using standard HTM date-time and scalar encoders~\cite{Purdy2016EncodingDF}. 

The SDR $a(x_t)$ produced by the SP is then sent to the SM, which is responsible of learning and processing sequence patterns that are used for prediction, taking into account the data processed in the past. The output of the SM is a sparse vector $\pi(x_t)$ that represents a prediction of the next input.

In contrast with other statistical and machine learning algorithms, such as neural networks, HTM better matches the characteristics of cloud environments. In fact, other algorithms are often trained in batch mode, requiring the storage of a large set of sequences to optimize performance on that specific data~\cite{Cui:ContinuousLearning:NC:2016,Cui:HTMvsNeuralNetwork:IJCNN:2016}. HTM is more flexible, providing full online and real-time learning, jointly with quick adaptation to new patterns, without requiring to store large sets of sequences. 

A detailed presentation of HTM is available in~\cite{Hawkins:SequenceMemory:FNC:2016,Cui:ContinuousLearning:NC:2016}.

\section{Online Failure Prediction in the Cloud} \label{sec:prediction}


\textcolor{revAle}{We study the effectiveness of HTM in supporting  cloud failure prediction by using a KPI-driven online failure prediction approach, which analyzes the KPIs to detect the patterns of anomalous behaviors that are likely to result in failures at a later time.}
 KPI-driven online failure prediction is implemented by many techniques~\cite{Lin:PredictingNodeFailures:ESECFSE:2018,Mariani:PredictingFailures:JSS:2020,Kanoun:AnomalyDetectionVNF:ISSRE:2016} and can be considered as the most used approach to failure prediction in the Cloud.

\begin{figure}
 \begin{center}
   \includegraphics[width=1\columnwidth]{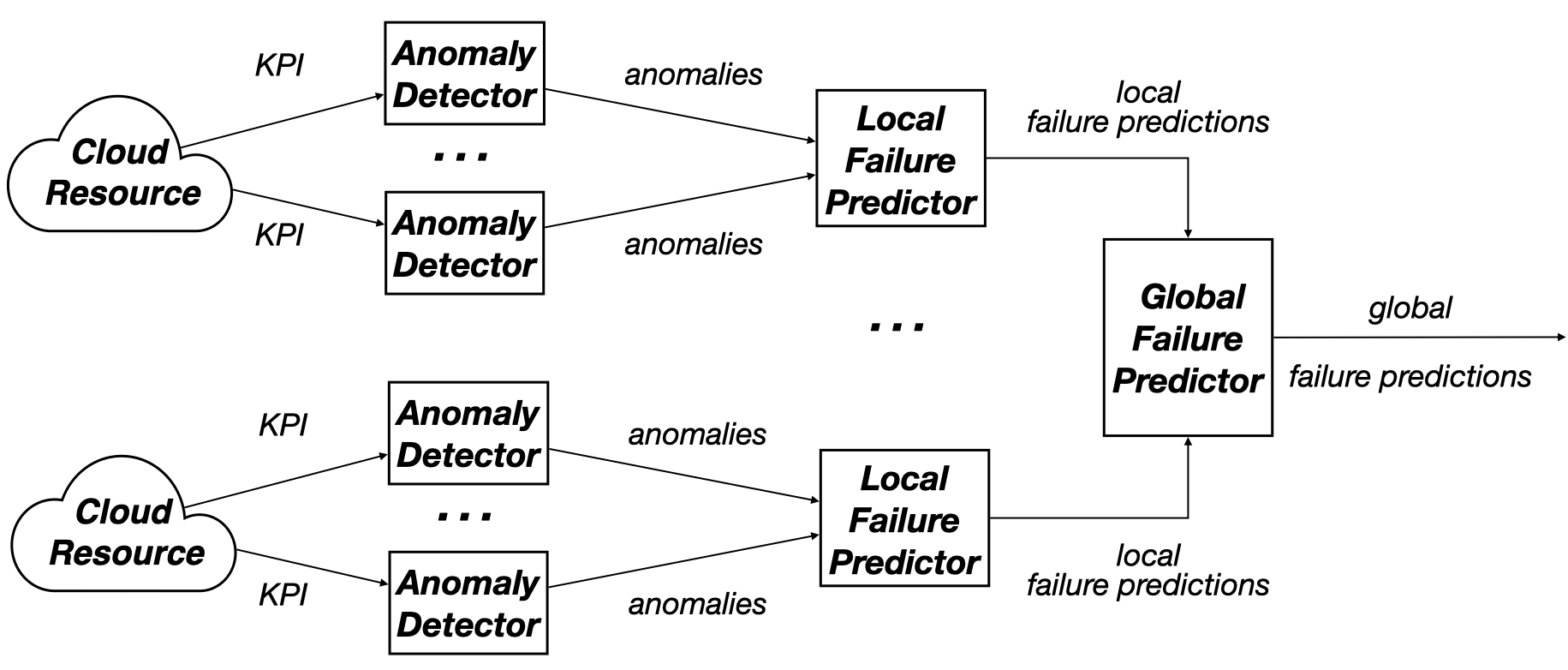}
   \caption{Online Failure Prediction}
 \label{fig:onlinePredicting}
 \end{center}
\vspace{-0.7cm}
 \end{figure}


The failure prediction system consists of three main types of components, as shown in Figure \ref{fig:onlinePredicting}: the \textit{Anomaly Detector}, which is responsible of identifying anomalous KPI values;  the \textit{Local Failure Predictor}, which is responsible of predicting failures based on the anomalies reported for the KPIs collected from a same cloud resource
; and the \textit{Global Failure Predictor}, which is responsible of ultimately predicting system failures by analyzing the resource-level failure predictions. The architecture includes one anomaly detector per KPI that is analyzed, one local failure predictor per cloud resource that is analyzed, and a single global failure predictor. \textcolor{rev}{Faulty executions of the system are not required to train the models.}

\subsection{Anomaly Detector}

The Anomaly Detector is the key component that can discriminate legal and anomalous values by analyzing each KPI separately, according to a learned predictive model. In this study, the anomaly detection algorithm employs Hierarchical Temporal Memory (HTM) as proposed by Ahmad et al. \cite{Ahmad:HTMAnomalyDetection:Neurocomputing:2017}. Note that HTM neural networks continuously learn and model the spatiotemporal characteristics of the inputs to predict the values of the inputs at the next time interval, and do not generate anomalies. However, they can be simply adapted to report anomalies, as proposed by Ahmad et al.~\cite{Ahmad:HTMAnomalyDetection:Neurocomputing:2017}.

Given a real-time KPI stream, where $x_t$ is a vector representing the state of the KPI at time $t$, HTM learns and predicts the temporal sequences of the values in such stream. Recall that the input $x_t$ is first semantically encoded into a form of sparse distributed representation (SDR), then normalized into a bit vector of fixed size and sparsity. The resulting vector $a(x_t)$ is used by HTM to produce a prediction in the form of another normalized sparse vector $\pi(x_t)$, which represents the prediction for the input at the next time interval $a(x_{t+1})$. 


HTM can be used to determine if $x_t$ is anomalous by comparing actual values to predictions made at the previous time interval $\pi(x_{t-1})$~\cite{Ahmad:HTMAnomalyDetection:Neurocomputing:2017}. Since $x_t$ and $\pi(x_{t-1})$ are bits vectors, a prediction error between 0 and 1 can be calculated depending on the \textit{``similarity''} between the actual and predicted bits, that is, the prediction error $S_t$ is inversely proportional to the number of common bits between the actual and predicted binary vectors. More rigorously:
\begin{equation*}
S_t=1-\frac{\pi(x_{t-1})\cdot a(x_t) }{\left | a(x_t) \right |}
\end{equation*}

where $\cdot$ indicates the multiplication operation between the bit vectors and  $| |$ denotes the number of elements in the vector.

The prediction error represents an instantaneous measure of the predictability of the current KPI stream. However, a threshold-based anomaly detection for this measurement could lead to a high number of false positives, especially in noisy systems. For this reason, an \emph{anomaly likelihood} $L_t$, which is a probabilistic metric that defines how anomalous the current state is based on the prediction history of the HTM model, is also computed~\cite{Ahmad:HTMAnomalyDetection:Neurocomputing:2017}. The history consists of a window of the last $W$ prediction errors. Assuming that the errors of the predictions have a rolling normal distribution, then the mean $\mu_t$ and the covariance $\sigma^{2}_t$ are continuously updated as follows:

\begin{align*}
    \mu_t=\frac{\sum_{i=0}^{i=W-1}S_{t-i}}{W} \qquad\quad \sigma_t^2=\frac{\sum_{i=0}^{i=W-1}(S_{t-i}-\mu_t)}{W-1}  
\end{align*}

%

A threshold to the Gaussian tail probability (Q-function) is applied to decide if an input is anomalous:

\begin{equation*}
L_t=1-Q(\frac{\tilde{\mu}_t-\mu_t}{\sigma_t})
\end{equation*}

where $\tilde{\mu}_t$ is a short term moving mean value computed as $\mu_t$ but on a smaller window.
%
The threshold for $L_t$ is based on a user-defined parameter $\epsilon$: an anomaly is reported if $L_t \leq 1-\epsilon$. 

\subsection{Local Failure Predictor}
The Local Failure Predictor analyzes anomalies in the KPIs collected from a same cloud resource to determine if they are symptoms of a failure or are anomalous but legal behaviors. The Local Failure Predictor is obtained by training a model with the outputs of the anomaly detector obtained during normal executions (note that several anomalous behaviors are usually generated also during normal executions). Deviations from normal behaviors, that is sets of anomalies different from the ones reported during normal executions, are reported as symptoms of a possible failure in the cloud resource. The possibility to train models only using normal executions is of fundamental importance for cloud failure prediction techniques, since there are many different failure situations that cannot be anticipated, and thus a fairly complete set of samples of failures is rarely available for a system.

In this experiment, we train a one-class Support Vector Machine (SVM) as Local Failure Predictor 
since it can be trained with data that is assumed to belong to a same class, which, in this case, can be easily obtained from the normal execution of the software. \textcolor{rev}{The SVM need to be updated only if the list of the KPIs monitored in a resource changes.} 
To  consider stable predictions only, the Local Failure Predictor is instructed to report a failure prediction only after a local prediction is confirmed for $n$ consecutive instants of time, where $n$ is a user-defined threshold. 

\subsection{Global Failure Predictor}
While the Local Failure Predictor performs a per-resource analysis, the goal of the Global Failure Predictor is to perform an ensemble analysis and produce a single failure prediction for the entire system. In fact, there might be anomalous behaviors limited to one resource that are handled by the resource, not propagating to other cloud resources and then to the rest of the system. The role of the global failure predictor is to determine when failures predicted locally may result in a system failure.

At each time interval $t$, the Global Failure Predictor accesses the predicted state of each resource and uses this information to confirm whether a local failure prediction is going to affect the whole system.
Similarly to Sauvanaud et al. \cite{Kanoun:AnomalyDetectionVNF:ISSRE:2016}, we investigate two strategies: (i) single-resource global prediction and (ii) vote-based global prediction. 

In the \emph{single resource prediction}, a system failure is predicted after $x$ consecutive local predictions of failure for the same resource, where $x$ is a user-configurable threshold. This case corresponds to a policy that assumes that local failures are likely to propagate at the system level. In the \emph{vote-based prediction}, instead, a failure is confirmed by the Global Failure Predictor when there are $y$ consecutive failure predictions on at least half of the resources, where $y$ is user-defined parameter. The idea here is that the system can tolerate local failures and only when multiple cloud resources are likely to fail, a system-level failure is likely experienced. 

\section{Methodology} \label{sec:methodology}

\textcolor{revAle}{We empirically assess HTM as online anomaly detector supporting prediction tools for cloud systems failures. In this section, we introduce the research questions that we studied, the telco cloud-based system we used as testbed for the assessment, and the fault seeding strategy that we adopted to collect data about failures caused by different types of faults.}

\subsection{Research Questions}
To assess HTM in the context of failure prediction, we investigate the following research questions:

\textcolor{revAle}{ \textit{RQ1: Can an HTM-based anomaly detector support a failure prediction system in accurately predicting failures?}}
We executed the prediction system with failure-free executions and with different failure types and failure patterns, and measured its ability to predict failures.

\textcolor{revAle}{ \textit{RQ2: How early can failures be predicted? }}
We studied how early failures are predicted, for different types of failures. 
\smallskip

These research questions are addressed by analyzing $72$ parameter configurations of the prediction system. \textcolor{rev}{The studied configurations are obtained by combining the following parameters with the specified values}:

\begin{itemize}[leftmargin=*]

\item $\epsilon$ is the anomaly likelihood threshold used for KPI anomaly detection. We considered $\epsilon = 0.8$, $0.85$, $0.9$ and $0.95$.
\item $n$ is the local prediction threshold of a possible failure on a single resource. We considered $n = 1$, and $2$.
\item $x$ and $y$ are thresholds used in single-resource and vote-based global prediction, respectively. We considered $x$ between $1$ and $6$, and $y$ between $1$ and $3$.
\end{itemize}



To answer RQ1, we evaluate the quality of the models using the standard metrics precision, recall and F-measure \cite{Salfner:OnlineFailurePredictionSurvey:2010}, so as to assess effectiveness in predicting failures. We address the research question RQ2 by calculating the time between a global failure prediction and its occurrence. 

\subsection{Testbed}
We perform the evaluation on a cloud environment running Clearwater, which is an open source implementation of an IP Multimedia Subsystem (IMS) that provides IP-based voice, video and message services \cite{clearwater}. Clearwater is a meaningful subject for this study since it represents the case of a ever-running cloud system that must operate reliably to guarantee IP-based communication. Failures must be indeed predicted and handled before they cause any service interruption. 


We used the standard installation of Clearwater, which consists of six components, each one running on a  different virtual machine (VM) configured with 2 vCPUs, 2GB of RAM,  20GB of hard disk space, and Ubuntu 12.04 LTS. In this evaluation, we detect anomalies from a total of 150 KPIs collected from all the components of Clearwater. The monitored KPIs include CPU and memory consumption, network usage, and many more. The detailed list of monitored KPIs is available in our online appendix to this paper~\cite{onlineAppendix}.
 
 \subsection{Experimental setup}
 We evaluate the quality of HTM by studying how it can support automatic failure prediction. Specifically, we first run Clearwater without producing any failure to train both the HTM models and the local classifier. In particular, data collected from a first week of failure-free executions is exploited to generate the HTM models, and data from a second week of failure-free executions is processed to train the local predictor. 
Note that failure-free executions are not free of anomalies, since services may operate in an unpredictable way, despite not generating any failure. To generate the traffic for the running system, we generated a number of users and calls according to weekly and daily patterns (e.g., more users on workdays, fewer users at night, pick time at 9am and 7pm), including a certain degree of randomness, as done in similar studies that used Clearwater as subject system~\cite{Mariani:PredictingFailures:JSS:2020}. 
 
To evaluate failure prediction, we inject the following four types of faults, one at time: CPU hogs, memory leaks, packet loss faults and excessive workload conditions as defined in \cite{ChaosMonkey:2021,Sharma:CloudPD:2013:DSN}. We activate the injected faults according to three activation patterns: (i) the fault is triggered with a same frequency over time; (ii) the fault is activated with a frequency that increases exponentially, resulting in a shorter time to failure; (iii) the fault is activated randomly over time. Overall, we experience 12 faults: 4 fault types times 3 activation patterns. We also assess failure prediction with 12 normal executions, not included in the training set, where the system was running under its normal operating conditions without faults or abnormal workloads, to assess the capability to not generate spurious predictions. 
\section{Results} \label{sec:results}
In this section we discuss the results obtained for the two research questions introduced in the previous section.

\subsection*{RQ1: Prediction Effectiveness}
We evaluated the capability of HTM to support failure prediction by studying the effectiveness of the vote-based and single-resource prediction strategies. In the experiments, failures correspond to either system crashes or a call success rate lower than $60\%$. We report the detailed results obtained for each configuration in the online appendix~\cite{onlineAppendix}. Here we discuss the most relevant results.

\begin{figure}[t]
\centering
\subfigure{
\includegraphics[width=.49\textwidth]{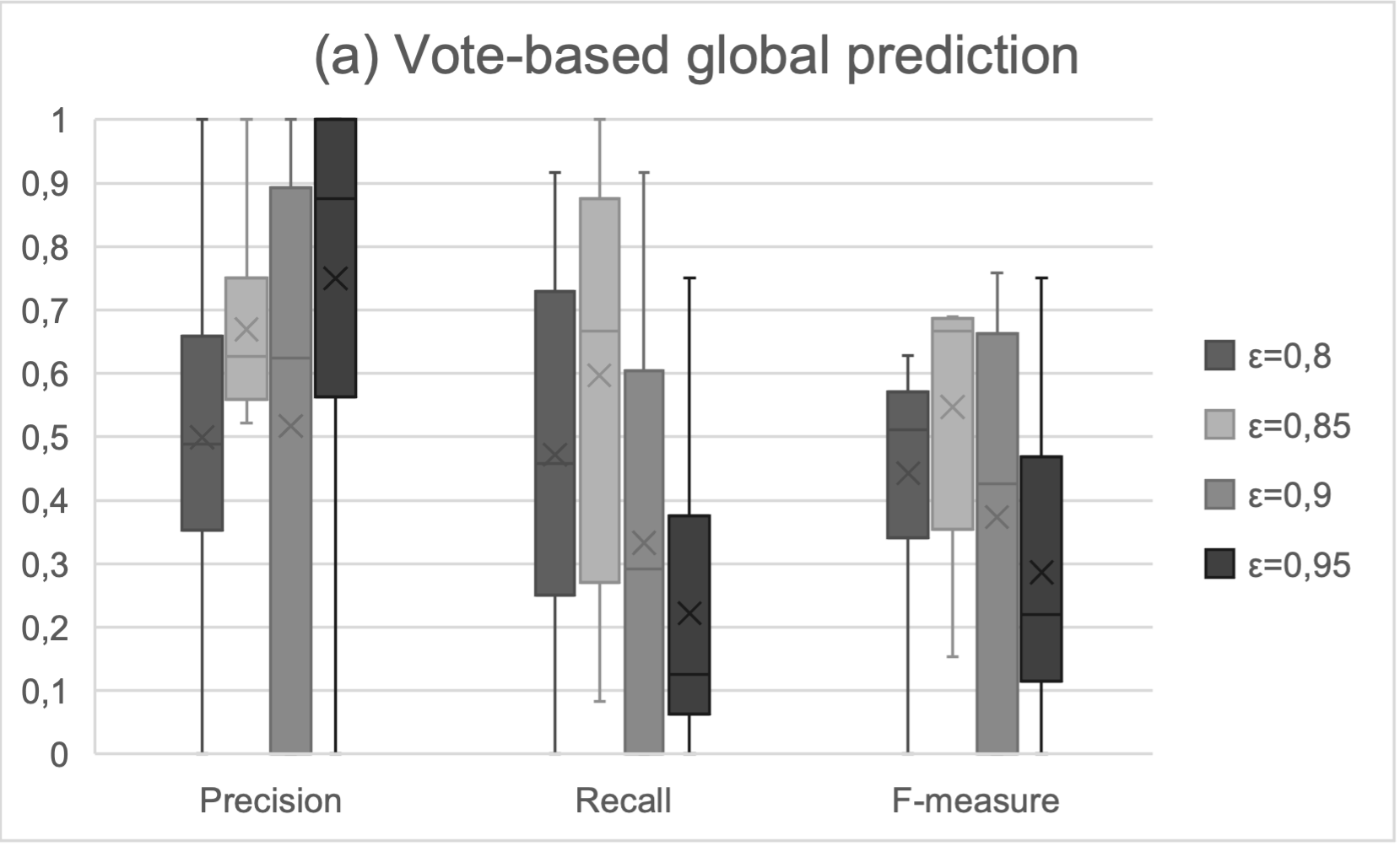}
}
\subfigure{
\includegraphics[width=.49\textwidth]{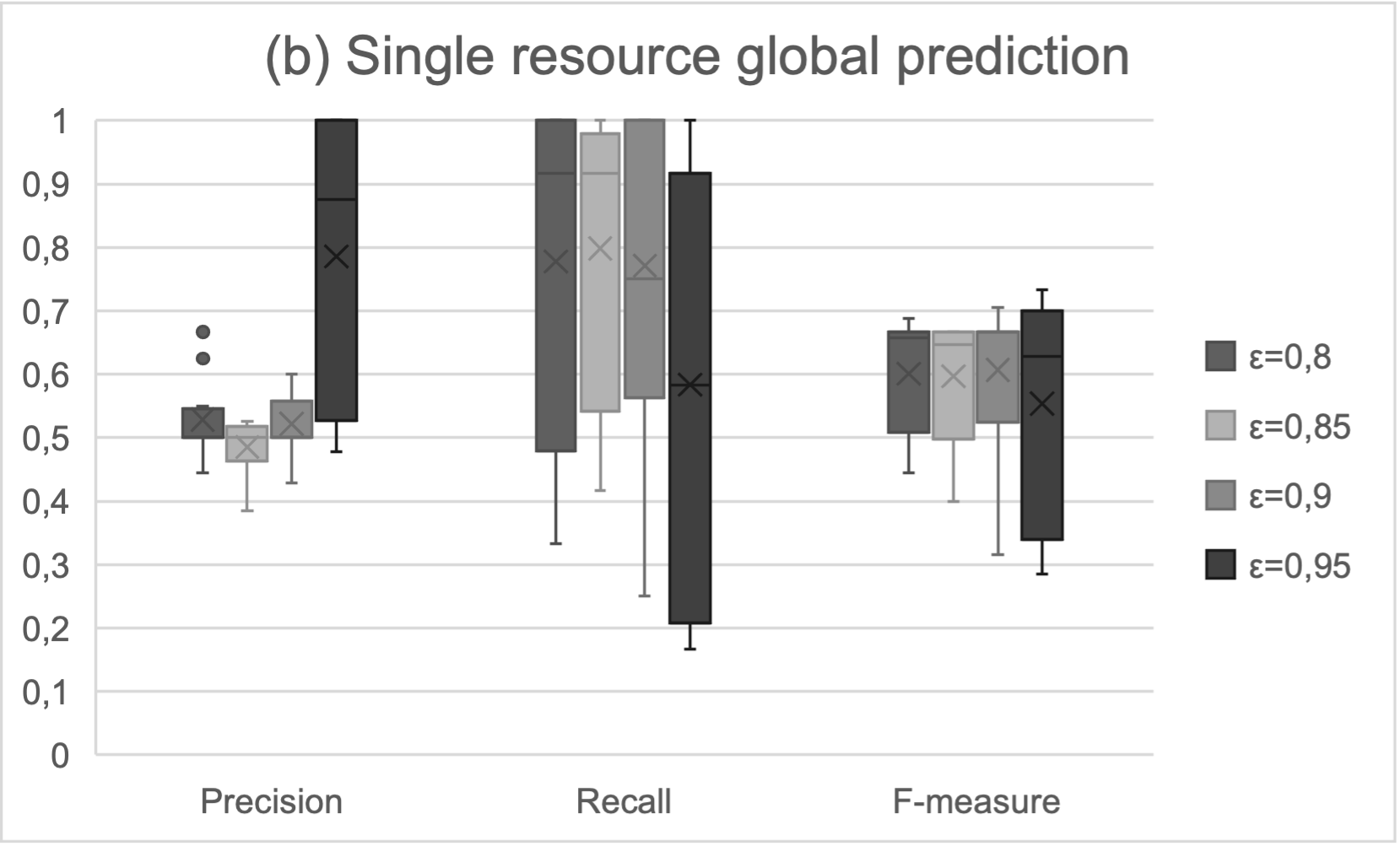}
}


\caption{Performance of global predictors}
\label{fig:preformancePredictors}
\vspace{-0.5cm}
\end{figure}

Figure~\ref{fig:preformancePredictors} (a) shows the precision, recall, and F-measure values of vote-based global prediction for the various configurations, aggregated according to the value of $\epsilon$. As expected precision increases while recall decreases for increasing values of $\epsilon$. The best performing configuration in terms of F-measure is the one with $\epsilon=0.9$, $n=1$ and $y=1$, achieving F-measure $= 0.76$, precision $= 0.65$, and recall $= 0.92$. This is also the configuration providing the highest recall. 
In some cases, engineers may want to optimize precision (e.g., to avoid false alarms). To this end, the configuration that provides the highest precision (i.e., precision $= 0.75$) with a recall greater than $0.5$ (i.e., recall $=0.75$) is $\epsilon=0.95$, $n=1$ and $y=1$.

Figure~\ref{fig:preformancePredictors} (b) shows the prediction, recall, and F-measure values of the single-resource global prediction strategy for the various configurations, aggregated according to the value of $\epsilon$. The trends of the precision, recall and F-measure resemble the ones of the vote-based global prediction, although differences between configurations are smaller. The configurations that provide the highest F-measure are the one with $\epsilon=0.95$, $n=1$, and $x=3$, and the one with $\epsilon=0.95$, $n=2$, and $x=2$, which both obtain F-measure $= 0.73$, precision $= 0.61$, and a recall $= 0.92$. These are also the configurations that provide the highest recall.
The configurations that provides the highest precision (i.e., precision $=0.88$) with recall greater than $0.5$ (i.e., recall $=0.58$) are the one with $\epsilon=0.95$, $n=1$ and $x=4$ and the one with  $\epsilon=0.95$, $n=2$ and $x=3$.

\textcolor{revAle}{Overall, these results show that HTM can be feasibly employed in supporting failure prediction}, especially if high recall values are needed (i.e., it is important to predict most of the failures at the cost of experiencing some false alarms). In this case, the vote-based strategy might be preferred to the single-resource strategy, since it achieves higher f-measure, with more stable results across configurations. 

Although it might be hard to achieve nearly perfect precision, HTM can still be employed when high precision is the priority. In such a case, the single-resource global prediction strategy should be preferred, since it reached a precision of $0.88$, without dropping recall below $0.5$.  

Interesting, a value of $n=1$ generates the best results for both strategies, suggesting that local predictions should be fired immediately, with the global predictor taking decisions considered the stability and spreading of failure predictions. In particular, the vote-based strategy performs best when firing a global failure prediction as soon as the majority of the cloud resources are likely to fail ($y=1$ in the best configurations). While, the single-resource strategy performs best when waiting for enough consistent local predictions before issuing a global failure prediction ($x=3$ or $4$ in the best configurations). 

Table~\ref{tab:faultAnalysis} shows recall per type of fault. The most difficult type of failure to predict is the ones caused by excessive workload, since an overloading number of requests generates behaviors quite similar to failure-free executions. The other failure types have similar recall, with cpu hogs, due to their impact on running services, being simpler to detect than other faults. 

\begin{table}[]
\centering
\caption{Fault Analysis}
\label{tab:faultAnalysis}
\begin{tabular}{l|c|c}
\textbf{Fault}     & \textbf{Recall} & \textbf{Median Prediction Time} \\ \hline 
CPU Hog            & 0.74            & 134 mins                        \\ \hline
Memory Leak        & 0.6             & 54 mins                         \\ \hline
Network Loss       & 0.65            & 51 mins                         \\ \hline
Excessive Workload & 0.52            & 15 mins                        
\end{tabular}
\vspace{-0.5cm}
\end{table}

Compared to studies where failures are used in the training phase~\cite{Kanoun:AnomalyDetectionVNF:ISSRE:2016,Mariani:PredictingFailures:JSS:2020}, HTM achieved slightly lower effectiveness. Although results obtained in different experiments are not directly comparable, a gap was expected. However, the use of failure-free executions only and the continuous training of HTM-based anomaly detectors represent important features that prevent engineers from spending a significant, and sometime impractical, effort in simulating hundreds of failures to train prediction models.



\subsection*{RQ2: Prediction timeliness}
\begin{figure}
 \begin{center}
   \includegraphics[width=1\columnwidth]{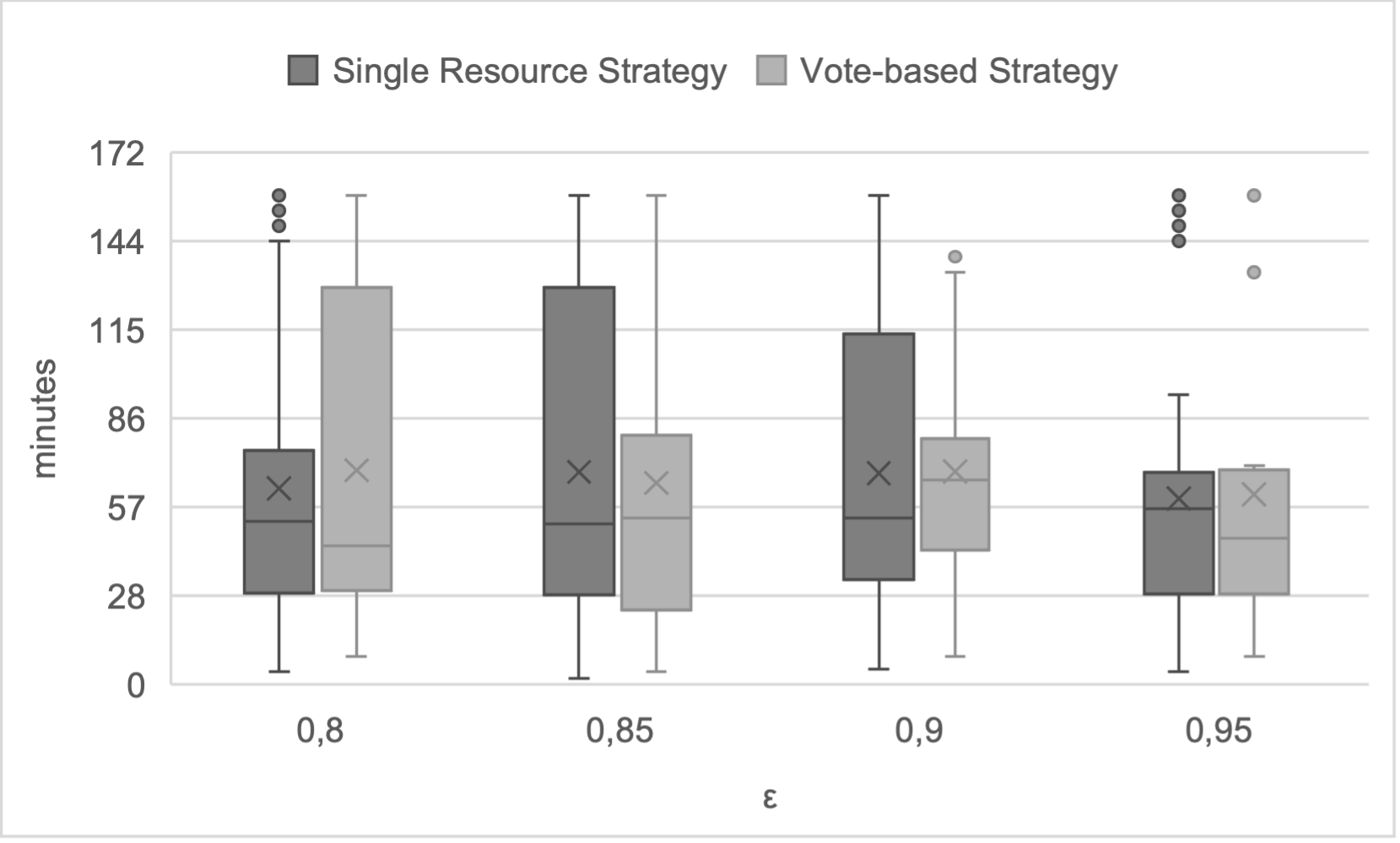}
   \caption{Prediction timeliness}
 \label{fig:predictionTime}
 \end{center}
 \vspace{-0.7cm}
 \end{figure}

RQ2 studies the capability to early predict failures. To evaluate the timeliness of the predictions, we measure the time difference between the prediction time and the failure time. This measure approximatively captures the time available to react to a failure prediction before the failure happens. Figure~\ref{fig:predictionTime} shows the results aggregated according to the value of $\epsilon$.

The vote-based global failure prediction was able to produce a failure prediction between 4 and 159 minutes before the failure occurred, with a median time of 54 minutes. The configuration with the best f-measure predicted the failure with a median time of 64 minutes.

The single-resource global failure prediction was able to produce a failure prediction between 2 and 159 minutes before the failure occurred, with a median time of 53 minutes. The configurations with the best f-measure predicted the failure with a median time of 57 minutes.

These time figures indeed allow the activation of automatic workaround, such as cloning or migrating services, and in the vast majority of the cases they are also early enough to allow for manual intervention.

Table~\ref{tab:faultAnalysis} reports the median failure prediction time per type of fault. Failures caused by excessive workloads have the smallest prediction time, since these failures are usually recognized only once the system is already congested.  The failures caused by CPU hogs are on the contrary easy to anticipate, due to their quickly recognizable impact on the running services.

\section{Related Work}\label{sec:related_work}
The work presented in this paper relates to research in two main areas: research about applications of HTM and research about prediction of failures in cloud systems.

\subsection{Applications of HTM}
HTM networks have been used to address various learning tasks in different domains. In particular, they are involved in providing real-time predictions in contexts where continuous and unpredictable changes affect the  input data.
For example, HTM has been employed for patient health monitoring and human-machine gesture interactions prediction~\cite{HTMMedicalStreams, HTMGazeGestureRecognition}, network intrusion detection~\cite{khangamwa2010detecting, zhang2020online}, phone network intention estimation~\cite{HTMCellularPhoneIntention}, short-term prediction of traffic flows~\cite{HTMTrafficFlow} and detection of anomalies in crowd movements
~\cite{Bamaqa:HTMCrowd:ICCBDC:20}.
The results reported in these works motivated the investigation of HTM applied to cloud failures prediction.


Ahmad et al.~\cite{Ahmad:HTMAnomalyDetection:Neurocomputing:2017} recently described how to use HTM to perform unsupervised real-time anomaly detection on streaming data. Their algorithm is capable of detecting spatial and temporal anomalies in predictable and noisy domains without look ahead and supervision. Although the original paper does not suggest any specific application scenario, the suggested anomaly detection schema well fits the domain of cloud failures prediction, and thus we used it in our experiments. 


One of the closest applications of HTM is the one by Rodriguez et al.~\cite{HTMScientificWorkflows}, who propose to use  HTM to detect anomalous resource consumption caused by the execution of scientific workflows. Differently for our study, anomalies are used to trigger resource scheduling and not to predict failures.


\subsection{Failure Prediction in Cloud Environments}

Several studies propose to predict failures using supervised machine learning algorithms that generate system-level models, that is, a model that captures the behavior of the entire system. For example, Lin et al.~\cite{Lin:PredictingNodeFailures:ESECFSE:2018} used an ensemble of supervised machine learning models  
to predict failures in cloud systems. 
Mariani et al.~\cite{Mariani:PredictingFailures:JSS:2020} combined anomaly-based and signature-based techniques for predicting failures in multi-tier distributed systems. Supervised learning is often impractical, since it needs generating many sample failures, while changes to the system require retraining the models. On the contrary, HTM offers an unsupervised continuously learning schema for streamed data that can represent a better choice for cloud failure prediction, as discussed in this paper. 




The failure prediction architecture considered in this paper has been exploited also by others. In particular, Sauvanaud et al.~\cite{Kanoun:AnomalyDetectionVNF:ISSRE:2016} proposed an approach for anomaly detection running both per-VM and system-wide analyses. Although they studied a different combination of techniques, it is interesting that both in their work and in the study presented in this paper system-wide (vote-based) analysis performed slightly better than the single-resource analysis, suggesting that a degree of global reasoning on the behavior of the system is often necessary to generate effective predictions. 

Related to our work, Mobilio et al.~\cite{Mobilio:ADaaS:ISSREW:2019} studied how to dynamically deploy and undeploy lightweight anomaly detectors in cloud systems. In the context of their evaluation, they considered HTM among the set of the anomaly detectors that can be deployed. Their approach only delivers preliminary findings about the possibility to use HTM as an anomaly detector for the cloud. This study strengthen this evidence, providing quantitative evidence on a larger scale.


Finally, a different line of research studies how to predict incidents using texts, issue reports, and statistical data~\cite{Zhao:RealTimePrediction:ESECFSE:2020,Gu:IncidentIdentification:ESECFSE:2020}. Differently, in this paper we studied the challenge of anticipating failures based on the KPIs collected online. 


\section{Conclusions}\label{sec:conclusions}

Predicting failures in cloud systems is a challenging problem that requires practical approaches to be solved effectively. Supervised learning schema that require retraining the models and exploit samples collected both during normal executions and during failures do not adapt well to this context. 

HTM is an interesting alternative offering an unsupervised continuously learning schema designed to work with streamed data, as the one collected when monitoring cloud systems. In this context, this paper offers a first systematic assessment of HTM, generating initial evidence that HTM can be a practical and effective option for \textcolor{revAle}{supporting} cloud failures prediction.

\textcolor{rev}{In this paper, we reported on our experience with a cloud native IMS. Future work concerns with validating cloud failure-prediction based on HTM in additional real-world systems.} 


\bibliographystyle{IEEEtran}
\bibliography{main}

\end{document}